\documentclass[letterpaper, 10 pt, conference]{ieeeconf}  % Comment this line out if you need a4paper

\IEEEoverridecommandlockouts                              % This command is only needed if 
\usepackage{graphicx} % for pdf, bitmapped graphics files
\usepackage[utf8]{inputenc}
\usepackage{amsmath,amsfonts,booktabs,cite} % general useful stuff
\usepackage{gensymb} % for \degree
\usepackage{siunitx,textcomp} % for the degree symbol
\usepackage[hidelinks]{hyperref} % to have hyperlinks and for the \ref macros
\usepackage{cleveref} % for easier referencing
\usepackage{subcaption}
\usepackage{tabularx}
\usepackage{makecell}
\usepackage{pbox}
\let\labelindent\relax
\usepackage[inline]{enumitem} % inline enumerate
\usepackage[nolist]{acronym} % acronyms by the \ac{label} command
\usepackage{multirow}
\usepackage[ruled,vlined]{algorithm2e}

\usepackage[font=small]{caption}

\def\figref#1{Fig.~\ref{#1}}

\def\eqref#1{Eq.~(\ref{#1})}

\newcommand\etal{~\emph{et al. }}

\newsavebox{\twosubbox}

\crefname{algocf}{alg.}{algs.}
\Crefname{algocf}{Algorithm}{Algorithms}

\title{\LARGE \bf
Fruit Mapping with Shape Completion for\\Autonomous Crop Monitoring
}

\author{Salih Marangoz \and  Tobias Zaenker \and  Rohit Menon \and Maren Bennewitz% <-this % stops a space
\thanks{This work was funded by the Deutsche Forschungsgemeinschaft (DFG, German Research Foundation) under Germany’s Excellence Strategy – EXC 2070 – 390732324. All authors are with the Humanoid Robots Lab, University of Bonn, Germany.}}

\begin{document}

\maketitle
\thispagestyle{empty} 
\pagestyle{empty}

\begin{abstract} 

Autonomous crop monitoring is a difficult task due to the complex structure of plants.
Occlusions from leaves can make it impossible to obtain complete views about all fruits of, e.g., pepper plants.
Therefore, accurately estimating the shape and volume of fruits from partial information is crucial to enable further advanced automation tasks such as yield estimation and automated fruit picking.
In this paper, we present an approach for mapping fruits on plants and estimating their shape by matching superellipsoids.
Our system segments fruits in images and uses their masks to generate point clouds of the fruits.
To combine sequences of acquired point clouds, we utilize a real-time 3D~mapping framework and build up a fruit map based on truncated signed distance fields.
We cluster fruits from this map and use optimized superellipsoids for matching to obtain accurate shape estimates.
In our experiments, we show in various simulated scenarios with a robotic arm equipped with an RGB-D camera that our approach can accurately estimate fruit volumes.
Additionally, we provide qualitative results of estimated fruit shapes from data recorded in a commercial glasshouse environment.

\end{abstract} 

\section{Introduction}
\label{sec:intro}

In the context of increasing demand for seasonal migrant farm labour in member European Union states and its supply side disruption due to the pandemic~\cite{granier2021migrant}, autonomous crop monitoring has gained further importance for plant phenotyping, yield monitoring, crop protection, and harvesting.
The availability of different sensors and systems has enabled crop monitoring at different levels from global low resolution satellite mapping~\cite{rembold2013using} to drone-based crop-level mapping~\cite{maimaitijiang2020crop,stefas2019vision} to individualized plant level mapping using mobile robots~\cite{qiao2005mapping} and mobile manipulators~\cite{shafiekhani2017vinobot}, with an increase in the level of detail and precision.
While global crop monitoring is necessary to make coarse yield predictions, fruit mapping is key to accurate yield monitoring and to performing automated precise interventions.

Fruit mapping consists of the localization of fruits and estimation of their shape.
With the advent of inexpensive RGB-D cameras, fusing instance masks from panoptic segmentation networks with depth information provides 3D information about the size and location of the fruits.
However, autonomous mapping of crops such as sweet peppers can be a challenging task since the plants have complex structures, with many leaves that may partially occlude fruits.
Therefore, obtaining accurate information about the location and shape of fruits can be difficult.
Even if observations from multiple views are acquired, fruits can often be only partially covered.

To enable shape prediction in these cases, we present a system to estimate the shape of both partially and fully observed fruits.
We build upon ideas of Lehnert\etal\cite{lehnert2016sweet} who proposed to match the shape of sweet peppers with superellipsoids.
Whereas their experiments were limited to controlled environments with single pepper plants, we integrated our superellipsoid matching system with our previously developed viewpoint planner~\cite{zaenker2020viewpoint}.
As a result, we are able to apply our fruit mapping system on data captured with an RGB-D camera on a robotic arm in various scenarios containing multiple plants with sweet peppers.

\begin{figure}
	\centering
	\includegraphics[width=\linewidth]{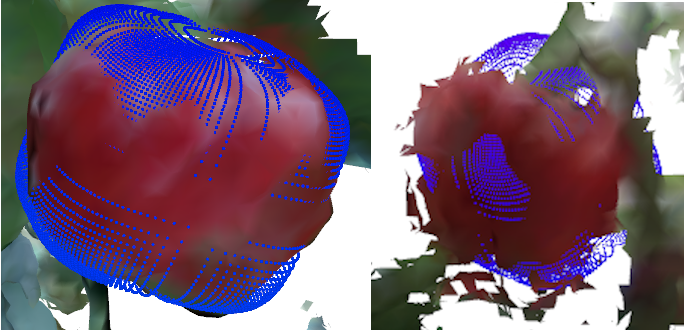}	
	\caption{Illustration of matched superellipsoids to fruits. Map generated from multiple views in a commercial glasshouse environment. Estimated superellipsoid surfaces are visualized as blue point clouds. While the left fruit has been observed from almost all directions, only half of the right fruit has been mapped. In both cases, our system was able to accurately predict their shape.
    }
	\label{fig:coverfig}
\end{figure}

\figref{fig:coverfig} shows an example application of our system using data of sweet pepper crops in a glasshouse.
As can be seen, fruit masks are detected in the RGB image to subsequently generate a fruit point cloud from depth data.
From a sequence of point clouds, we generate a fruit map in form of truncated signed distance fields~(TSDF) with the mapping framework voxblox~\cite{oleynikova2017voxblox}.
The surface cloud of this TSDF map is clustered to separate fruits, and superellipsoids are fitted by first estimating center and normals of the points and then minimizing the distance of detected points to the superellipsoid surface.

Our main contributions are the following: 
\begin{itemize}%[label=(\roman*)]
	\item A complete framework that accurately maps and estimates fruit shapes using multiple views,
	\item An implementation in ROS, providing easy integration with our previously developed ROI viewpoint planner,
	\item An evaluation of the accuracy of the estimated fruit volumes in various simulated scenarios with multiple crops, 
	\item Qualitative results on data recorded in a commercial glasshouse environment.
	
\end{itemize}

\section{Related Work}
\label{sec:related}

Several approaches exist that are based on reconstruction using 3D data collected from different viewpoints to obtain a full 3D~model, however, with no explicit modeling of shape.
For example, Wang\etal\cite{wang2020fruit} developed automatic fruit shape measurement based on 3D reconstruction, using a Kinect sensor and an electric turntable.
Similarly, Jadhav\etal\cite{jadhav2019volumetric} investigated the volumetric 3D reconstruction of different fruits based on voxel mapping using silhouette probabilities and inter-image homographies, with multiple RGB cameras and a checkerboard.
Complete reconstruction approaches are generally suitable for ground truth dataset generation or volumetric estimation in laboratory conditions.
However, the reconstruction methods are computationally expensive and are generally not suitable for resource-constrained, real-time agriculture automation tasks.
Moreover, occlusions due to the plant leaves or unreachability of the sensor to cover the complete view can lead to incomplete scanning of the fruits.

Viewpoint planning uses the hitherto accumulated information to plan the next best view for maximising the information gain.
As an example, Roy and Isler\cite{roy2017active} used a viewpoint planning approach to count apples in orchards, however, without any volumetric estimation.
In our previous work~\cite{zaenker2020viewpoint}, we formulated a sampling approach based on regions of interest~(ROI), i.e., fruits that outperforms those that do not consider ROIs, with respect to the number of detected fruits and the estimated volume.
However, the previous approach suffered from inaccuracies in fruit volume estimation as no shape estimation or modeling was carried out for predicting the unobserved fruit regions.

State-of-the-art shape completion or fruit modeling approaches do not use a viewpoint planner for maximizing the information gain, instead they typically attempt to predict fruit shape and pose using data from a single view or few views.
Kang\etal\cite{kang2020visual} used the DasNet deep learning network~\cite{kang2020fruit} for fruit recognition, and 3D Sphere Hough Transform to estimate the fruit pose and size, for autonomous apple harvesting.
Similarly, Lin \etal \cite{lin2019guava} demonstrated that sphere fitting works better than a bounding box method for pose estimation of guavas.
However, while spherical fitting provides acceptable results for autonomous harvesting, the reduction of different fruits to simple spheres is neither generalizable, nor reliable for localized yield predictions.
Ge\etal\cite{ge2020symmetry} formulated another approach based on shape from symmetry~\cite{thrun2005shape}, to generate and rate different hypotheses for the symmetrical plane and the mirrored point sets, for the partial view, to reconstruct the complete shape of strawberry fruits for fruit localization.
Furthermore, photogrammetry-based methods have been used for estimating real length and width after fitting ellipses for apples and mangoes, respectively, in the 2D plane~\cite{gongal2018apple,wang2017tree}.

The most closely related work is the approach by Lehnert\etal\cite{lehnert2016sweet} that fits superellipsoids to sweet peppers.
Here, point clouds from different viewpoints were merged into a single point cloud using different registration methods of which Kinect Fusion~(KinFu) was found to perform best at pose tracking.
After registration and segmentation of red sweet papers based on color using the HSV colour space, sweet pepper pose estimation is performed using constrained non-linear least squares optimization.
A superellipsoid model is then used for estimating parameters of the fruit shape and for obtaining the 6 DoF transform between the registered point cloud and the model.

KinFu can be used for realtime mapping of scenes while concurrently obtaining the camera pose by tracking the depth frames continuously as in~\cite{lehnert2016sweet}.
However, as the viewpoint planner used in our previous approach~\cite{zaenker2020viewpoint} plans discrete viewpoints for maximising information gain, KinFu cannot perform continuous tracking and registration which needs small movements in successive frames for the convergence of ICP alignment~\cite{pirovano2012kinfu}.
Oleynikova\etal\cite{oleynikova2017voxblox} developed voxblox to build TSDFs based map representations for local planning of Micro Aerial Vehicles.
Unlike KinFu, which requires a known map size, large amounts of memory, and GPU computational power, voxblox can dynamically grow maps as it adds point cloud data from different viewpoints.
As voxblox can run on a single CPU core with low memory requirements and still build TSDF maps faster than building occupancy maps, we replaced the 3D~occupancy maps in~\cite{zaenker2020ecmr} with TSDF maps.

In conclusion, most state-of-the-art approaches for fruit mapping fall into three categories -- naive complete reconstruction methods, viewpoint planning without any shape completion, and geometric shape completion methods.
Our approach combines a viewpoint planner with shape completion for mapping of fruits for estimation of count and volume, in the wild, with an improved TSDF based mapping.

\begin{figure*}[h] 	\centering 	\includegraphics[width=\linewidth]{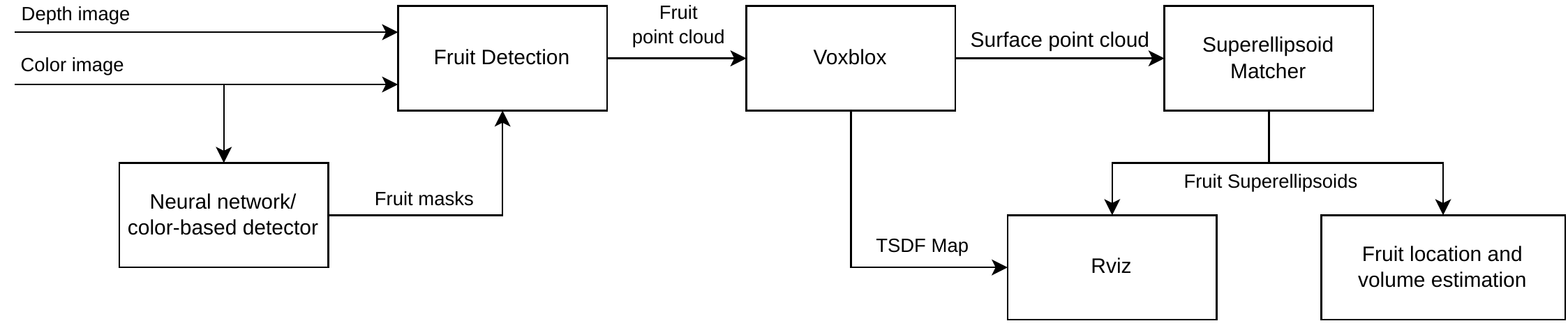} 	\caption{Overview of our system.
Color and depth images are converted into a fruit point cloud from generated fruit masks, which is processed by voxblox~\cite{oleynikova2017voxblox} into a truncated signed distance field~(TSDF) map.
The surface point cloud from this map is used to fit superellipsoids to individual fruits, which can be visualized together with the map, or used to estimate position and volume.} 
	\label{fig:system_overview}
\end{figure*} 

\section{System Overview}
\label{sec:approach}

Our approach aims at estimating the shape of detected fruits, which might be partially occluded.
We use our previously developed viewpoint planner~\cite{zaenker2020viewpoint} to move a camera placed on a robotic arm, which enables us to observe fruits from different perspectives.
In order to estimate the shape of detected fruits, we build a map of the environment.
Previously, we generated an octree with marked regions of interest~(ROIs), i.e., fruits, to guide the planner and estimate the fruit position and size.
While this enabled ROI targeted viewpoint planning and a rough estimation of fruit positions, due to the limited accuracy of the octree map, the shape of fruits could not be accurately determined.

For the new approach presented in this paper, we therefore build a separate map of the environment specifically for the fruits.
We use the masks of detected fruits %, either acquired through a neural network or color thresholding, 
to generate a separate fruit point cloud as input for the fruit mapping.
This point cloud is processed to remove outliers and forwarded to the voxblox mapping system~\cite{oleynikova2017voxblox}, which creates a TSDF map of the fruits.
Our superellipsoid matching system receives the surface point cloud generated from that map, clusters it into individual fruits, and optimizes the parameters of a superellipsoid to find the best fitting shape for the corresponding points.
The parameters of the superellipsoid can be used to compute the fruit position and volume.
Addionally, a fruit point cloud for visualization purposes is generated.
\figref{fig:system_overview} shows an overview of the approach described above.

\section{Fruit Mapping and Shape Estimation}

\subsection{Fruit Detection and Mapping}
\label{sec:detection_mapping}

In order to match superellipsoid to fruits, they have to be detected first.
Like in our previous work, we provide two methods to detect fruits in images.
The first one is a color thresholder for detecting red pixels, and is used in our simulated experiments with only red peppers.
For our real world data, we utilize a neural network trained on a dataset recorded at the University of Bonn~\cite{halstead2020fruit} to detect the fruit masks.
In both cases, we end with a set of pixels belonging to fruits.
We match those with the point cloud generated from the depth image of the RGB-D camera, and extract a cloud only containing the fruit points.
This cloud is used to generate the map used for our shape completion, while the complete cloud can be used to generate an additional map containing full plants.
Before that, outliers are removed using statistical outlier removal to counter depth noise.

For generating the map used to match superellipsoids to fruits, we utilize the mapping framework voxblox \cite{oleynikova2017voxblox}.
We configure voxblox with a TSDF voxel size of 0.004 m and 64 voxels per side.
We disable ICP tracking feature of voxblox and utilize position feedbacks from the robot arm.

\subsection{Clustering and Shape Estimation}
\label{sec:approach_sampling}

From the fruit TSDF map generated with voxblox, we extract the surface point cloud to match our fruit predictions.
In a first processing step, we extract clusters to identify individual fruits.
We utilize Euclidean cluster extraction with a cluster tolerance of 0.01 m, a minimum cluster size of 100 points and a maximum cluster size of 10000 points for that.
For each extracted cluster, we first estimate the normals of the points.
Taking into account the normals, a cluster center $w$ is estimated using a closed form least-squares solution for minimizing the sum of perpendicular distances from the estimated point to all the lines \cite{traa2013least}.
We also use L2 regularization to bias the solution towards a reference point (i.e., the mean of all points in a cluster) with the regularization value of $\lambda = 2.5$ \cite{traa2013least}.
This center is used as start point for the optimization of the superellipsoid.

As in \cite{lehnert2016sweet}, we represent superellipsoids with the following equation: 

\begin{equation}
	f(x, y, z) = \left[
	\left(\frac{x}{a}\right)^\frac{2}{\varepsilon_2} +
	\left(\frac{y}{b}\right)^\frac{2}{\varepsilon_2}
	\right]^\frac{\varepsilon_2}{\varepsilon_1} +
	\left(\frac{z}{c}\right)^\frac{2}{\varepsilon_1} = 1	
\end{equation}

The parameters $a$, $b$, $c$, $\varepsilon_1$, and $\varepsilon_2$, together with the translation $t_x$, $t_y$, $t_z$ and rotation $\phi$, $\theta$, $\psi$ make up 11 parameters to optimize.
We set the initial value of $t_x$, $t_y$, $t_z$ to the estimated center and the rotation to 0.
For the superellipsoid parameters, we start with initial values of $a=b=c=0.05\ m$ and $\varepsilon_1=\varepsilon_2=0.5$.

Compared to \cite{lehnert2016sweet}, we changed the cost function to achieve better fits.
Jakli{\v{c}} \etal \cite{jaklic2000superquadrics} describe approximated distance of a point to a superellipsoid as: 

\begin{equation}
	d = |r_0 - \beta r_0| = |r_0| \cdot |1-f^{-\frac{\varepsilon_1}{2}}(x_0, y_0, z_0)|
\end{equation}

where $r_0 = (x_0, y_0, z_0)$ represents the coordinates of a point with respect to the superellipsoid center.
$\beta$ is the scaling factor to project $r_0$ to the superellipsoid surface, and computed by solving $f(\beta x_0, \beta y_0, \beta z_0)$ for $\beta$.
To find the best fitting shape, we minimize the sum of $d$ over all points.
We get the following optimization task, with priors encouraging the superellipsoid center to be closer to the estimated center and the $a$, $b$, and $c$ parameters to be close to each other: 

\begin{equation}
	\min_{a, ..., \psi} \sum_i \biggl [ d_i^2 + \alpha ||t-w||_2^2 + \gamma \bigl((a-b)^2+(b-c)^2+(c-a)^2 \bigr) \biggr ]
\end{equation}

We apply the Levenberg–Marquardt algorithm to optimize the parameters.
The numerical computations are performed using the open source Ceres solver \cite{ceres-solver}.

We limit $a$, $b$, and $c$ between 0.02 and 0.15 m, and $\varepsilon_1$ and $\varepsilon_2$ between 0.3 and 0.9, to avoid unrealistically small, large or deformed superellipsoids that don't realistically represent a fruit.
We set regularization constants $\alpha = \gamma = 0.1$. We also limit the maximum number of iterations to 100 and discard optimizations that terminated without convergence.

\subsection{Shape Evaluation}
\label{sec:approach_evaluation}

Our node provides the results of the matched superellipsoids in multiple forms.
For visualization, surface point clouds are generated from the optimized parameters.
We provide a colored point cloud with different colors for each cluster, as well a labeled point cloud including the computed normals.
For further usage, the superellipsoid parameters and volume are published.
We compute the volume using the following formula, as described in \cite{jaklic2000superquadrics}: 

\begin{equation}
	V = 2abc\varepsilon_1\varepsilon_2B\left(\frac{\varepsilon_1}{2}+1, \varepsilon_1\right)B\left(\frac{\varepsilon_2}{2}, \frac{\varepsilon_2}{2}\right);
\end{equation}

where $B$ is the beta function: 

\begin{equation}
B(x, y) = 2\int_{0}^{\frac{\pi}{2}}
\sin^{2x-1}\phi cos^{2y-1} \phi d\phi = \frac{\Gamma(x)\Gamma(y)}{\Gamma(x + y)}
\end{equation}

For our simulated environments, we extract a point cloud from the mesh models of the fruits and optimize superellipsoid parameters to fit this point cloud for each fruit.
We read the position and orientation of the plants directly from the Gazebo simulator and can therefore generate a ground truth superellipsoid for each fruit in the scene.

We can match ground truth and detected ellipsoids by comparing the distance of the centers and the volumes.

\section{Experiments}
\label{sec:exp}

We carried out experiments with of our shape estimation method in a simulated environment for quantitative analysis and with data collected from the glasshouse to validate our approach in real world conditions.

For our evaluation in simulated scenarios, we used the Gazebo based simulation setup presented in \cite{zaenker2020viewpoint}, but with an additional new environment with more and different plants.
It simulates a UR5e arm from Universal Robots with an RGB-D camera placed at its flange. We configured the planner to move the arm to new views autonomously.

We placed the arm on a static pole for the first scenario, and for the other two, it was attached to a retractable pole hanging from the ceiling, to enable good coverage of the fruits.
The three scenarios are described with pictures in \Cref{fig:simulatedenv1,fig:simulatedenv2,fig:simulatedenv3}.

\subsection{Simulated Scenarios}

To evaluate the accuracy of the fruit shape estimation with our new method to our old approach, we compare the average estimated fruit center distance and volume accuracy of the detected fruits.
Additionally, we set a new restriction on the detected fruits -- they are counted only when the specified minimum accuracy criterion is met.
For the evaluation, we consider the number of fruits with minimum accuracy 0 (i.e. the detected volume must not be more than twice the ground truth volume) and minimum accuracy 0.5 (the detected volume must not be more than 50\% larger or smaller than the ground truth.)

This gives us the following metrics: \begin{itemize} 
	\item \textit{Detected fruits}: Number of found fruits 
that can be matched with ground truth fruits, which means that their center distance is less than~$20\,cm$, and their volume accuracy is greater than the specified minimum.
	\item \textit{Center distance}: Average distance of the fruit centers from the ground truth centers.
	\item \textit{Volume accuracy}: Average accuracy of the cluster volumes. 
	The accuracy is computed as follows: 
	\begin{equation}
		acc_{V} = 1 - \frac{|V_{det} - V_{gt}|}{V_{gt}}
	\end{equation}
	where $V_{det}$ is the detected volume and $V_{gt}$, the ground truth volume.
For detected volumes less than the ground truth volume, the $acc_{V}$ varies from 0 to 1, with 1 representing the best possible accuracy when they are equal.
On the other hand, for $V_{det}$ greater than $2V_{gt}$, the $acc_{V}$ turns negative.
	For our old method, the volumes are computed as bounding box volumes of the detected clusters.
	For the new method, we use the superellipsoid volume as described in \Cref{sec:approach_evaluation}.
\end{itemize} 

For each scenario, we executed 20 trials, each with a planning time of three minutes.

\begin{figure*} 	\centering 	\begin{subfigure}[b]{0.32\linewidth} 		\centering 		\includegraphics[width=\linewidth]{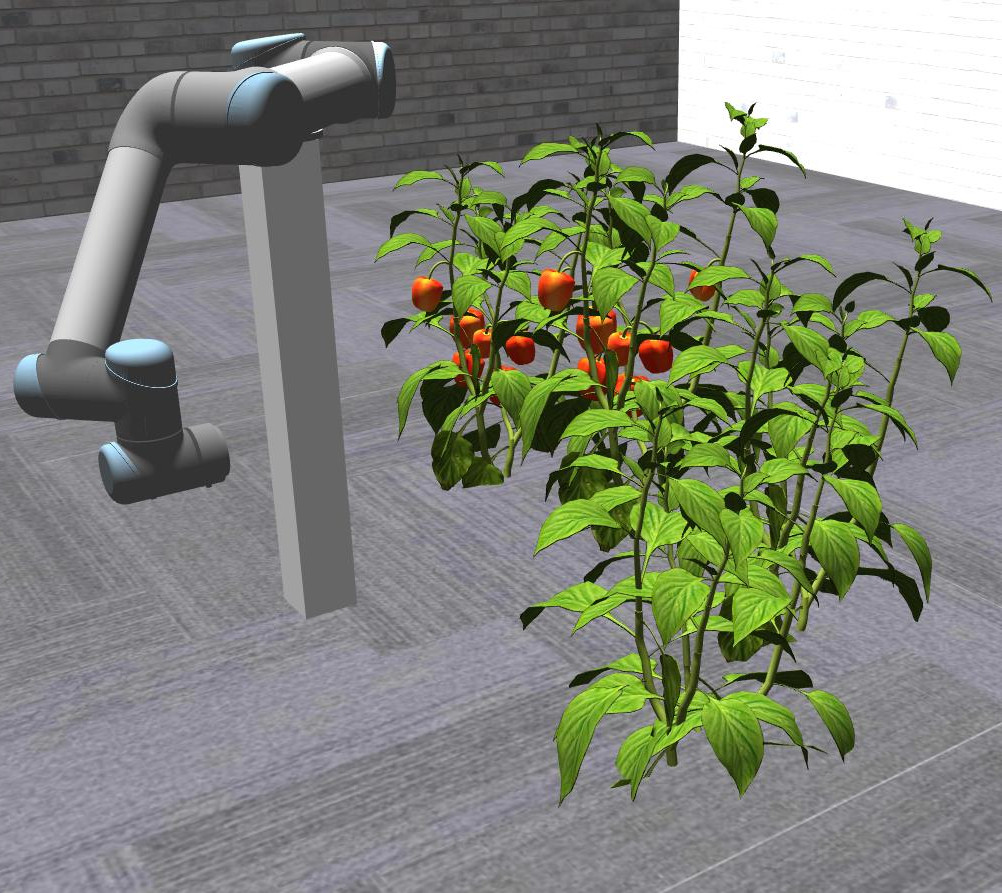} 		\caption{Scenario 1} 
		\label{fig:simulatedenv1}
	\end{subfigure} 	\begin{subfigure}[b]{0.32\linewidth} 		\centering 		\includegraphics[width=\linewidth]{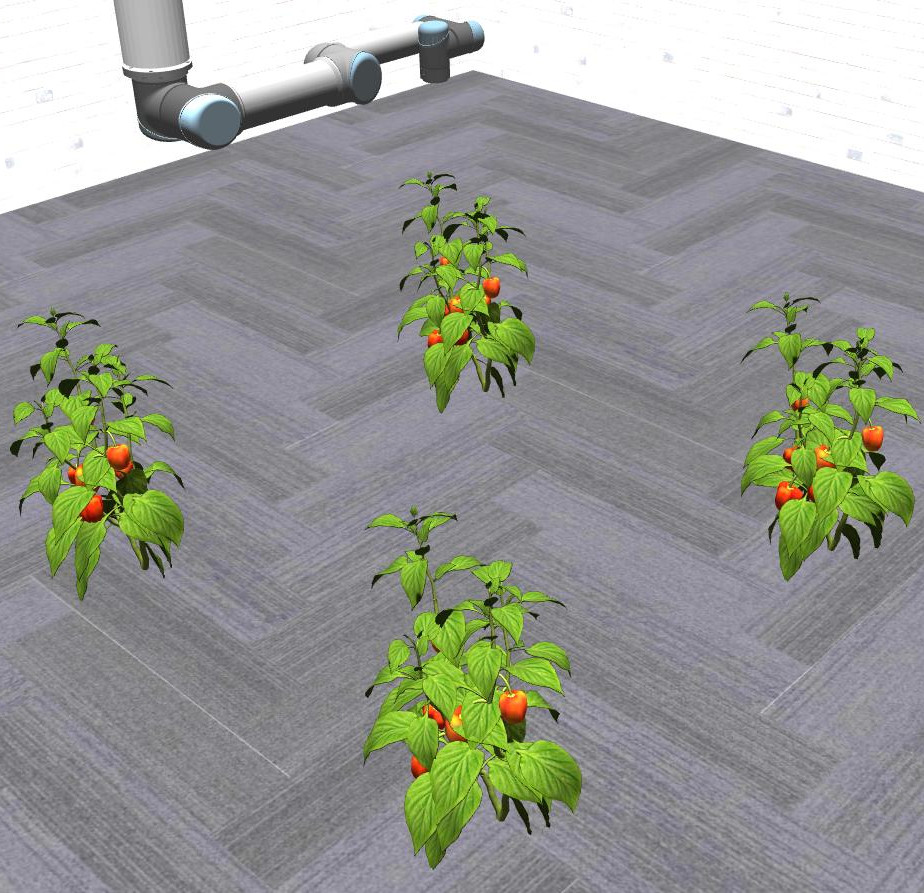} 		\caption{Scenario 2} 
		\label{fig:simulatedenv2}
	\end{subfigure} 	\begin{subfigure}[b]{0.32\linewidth} 		\centering 		\includegraphics[width=\linewidth]{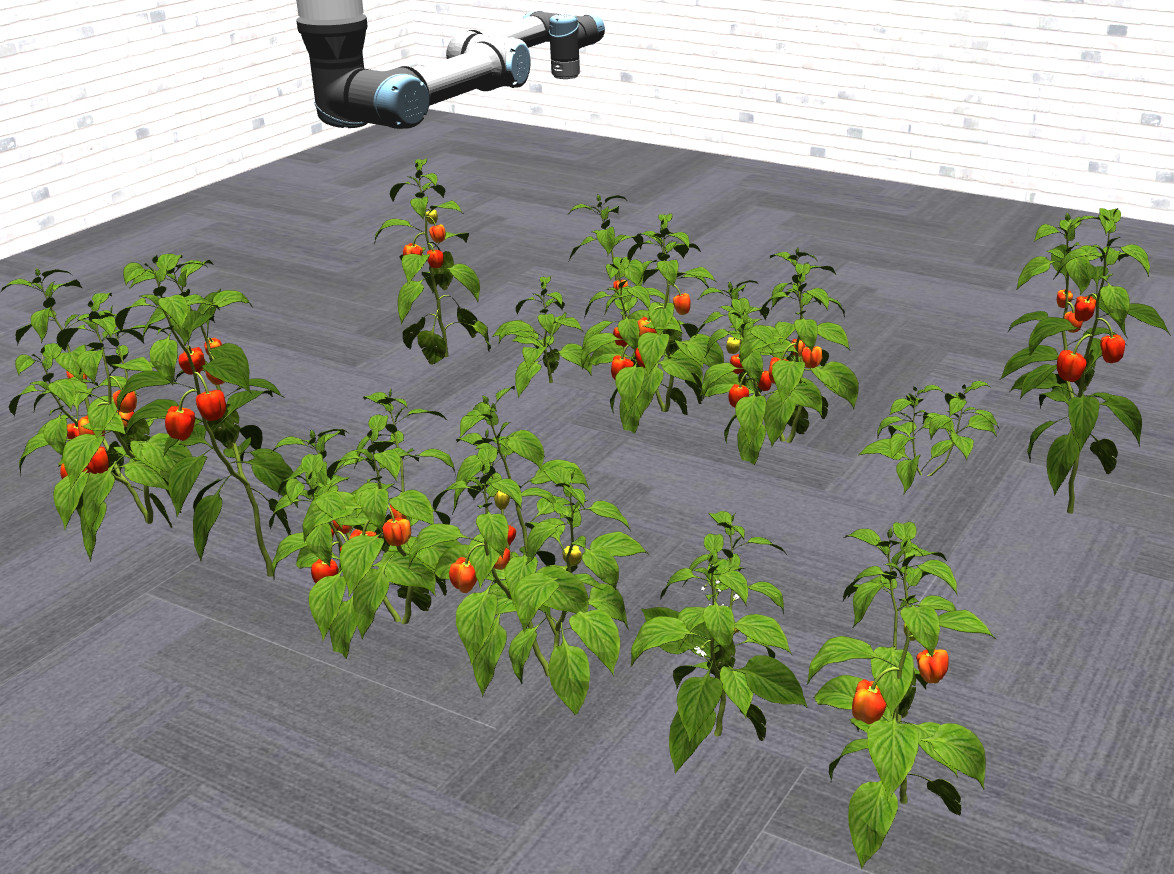} 		\caption{Scenario 3} 
		\label{fig:simulatedenv3}
	\end{subfigure} 	\begin{subfigure}[b]{0.32\linewidth} 		\centering 		\includegraphics[width=\linewidth]{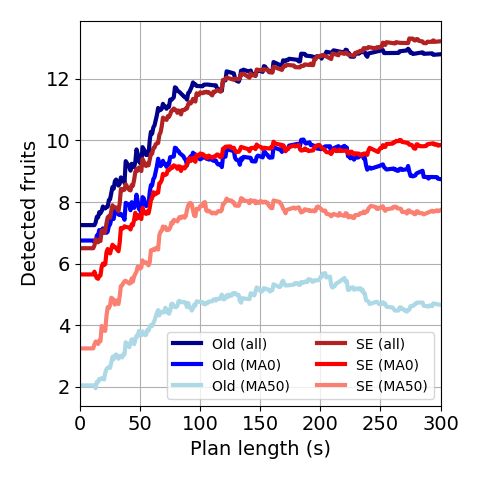} 		\caption{Fruits Scenario 1} 
		\label{fig:det_fruits_w14}
	\end{subfigure} 	\begin{subfigure}[b]{0.32\linewidth} 		\centering 		\includegraphics[width=\linewidth]{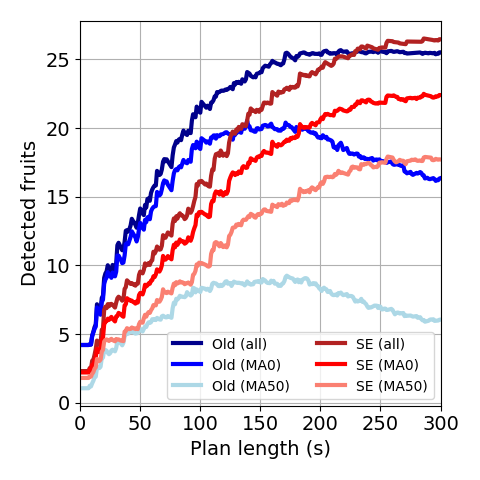} 		\caption{Fruits Scenario 2} 
		\label{fig:det_fruits_w19}
	\end{subfigure} 	\begin{subfigure}[b]{0.32\linewidth} 		\centering 		\includegraphics[width=\linewidth]{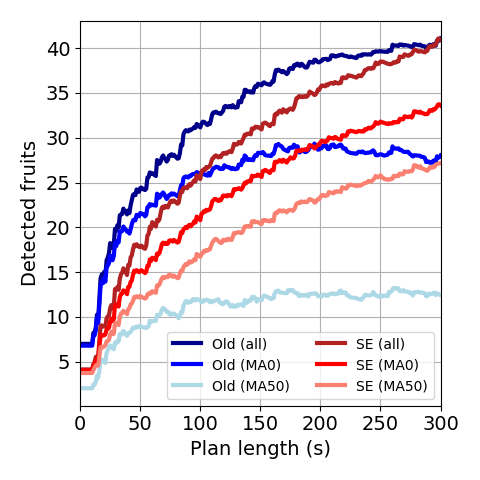} 		\caption{Fruits Scenario 3} 
		\label{fig:det_fruits_w22}
	\end{subfigure} \par\bigskip \begin{subfigure}[b]{0.32\linewidth} 		\centering 		\begin{tabular}{ c | c | c | c |} \cline{2-4} 		& Fruits & Center & Vol.
		\\ \hline\hline 
		\makecell[c]{Old\\(all)} & \makecell[r]{12.8\\ $\pm$ 0.9} & \makecell[r]{2.5\\ $\pm$ 0.5} & \makecell[r]{-0.12\\ $\pm$ 0.79} \\ \hline
		\makecell[c]{SE\\(all)} & \makecell[r]{13.2\\ $\pm$ 0.9} & \makecell[r]{2.9\\ $\pm$ 0.7} & \makecell[r]{0.26\\ $\pm$ 0.27} \\ \hline\hline
		\makecell[c]{Old\\(MA0)} & \makecell[r]{8.7\\ $\pm$ 2.6} & \makecell[r]{2.6\\ $\pm$ 1.1} & \makecell[r]{0.45\\ $\pm$ 0.12} \\ \hline
		\makecell[c]{SE\\(MA0)} & \makecell[r]{9.8\\ $\pm$ 1.9} & \makecell[r]{2.3\\ $\pm$ 0.8} & \makecell[r]{0.69\\ $\pm$ 0.06} \\ \hline\hline
		\makecell[c]{Old\\(MA50)} & \makecell[r]{4.7\\ $\pm$ 2.2} & \makecell[r]{1.6\\ $\pm$ 0.8} & \makecell[r]{0.65\\ $\pm$ 0.17} \\ \hline
		\makecell[c]{SE\\(MA50)} & \makecell[r]{7.7\\ $\pm$ 1.9} & \makecell[r]{1.9\\ $\pm$ 0.9} & \makecell[r]{0.81\\ $\pm$ 0.04} \\ \hline
	\end{tabular} \caption{Results Scenario 1} \label{tab:res_table_1} 	\end{subfigure} 	\begin{subfigure}[b]{0.32\linewidth} 		\centering 	\begin{tabular}{ c | c | c | c |} \cline{2-4} 		& Fruits & Center & Vol.
		\\ \hline\hline 
		\makecell[c]{Old\\(all)} & \makecell[r]{25.5\\ $\pm$ 1.0} & \makecell[r]{2.2\\ $\pm$ 0.3} & \makecell[r]{-0.05\\ $\pm$ 0.21} \\ \hline
		\makecell[c]{SE\\(all)} & \makecell[r]{26.5\\ $\pm$ 1.2} & \makecell[r]{1.8\\ $\pm$ 0.6} & \makecell[r]{0.35\\ $\pm$ 0.23} \\ \hline\hline
		\makecell[c]{Old\\(MA0)} & \makecell[r]{16.4\\ $\pm$ 3.4} & \makecell[r]{2.1\\ $\pm$ 0.5} & \makecell[r]{0.35\\ $\pm$ 0.06} \\ \hline
		\makecell[c]{SE\\(MA0)} & \makecell[r]{22.4\\ $\pm$ 2.4} & \makecell[r]{1.4\\ $\pm$ 0.6} & \makecell[r]{0.67\\ $\pm$ 0.04} \\ \hline\hline
		\makecell[c]{Old\\(MA50)} & \makecell[r]{6.0\\ $\pm$ 2.0} & \makecell[r]{1.8\\ $\pm$ 0.4} & \makecell[r]{0.66\\ $\pm$ 0.07} \\ \hline
		\makecell[c]{SE\\(MA50)} & \makecell[r]{17.7\\ $\pm$ 2.1} & \makecell[r]{1.2\\ $\pm$ 0.4} & \makecell[r]{0.78\\ $\pm$ 0.03} \\ \hline
	\end{tabular} \caption{Results Scenario 2} \label{tab:res_table_2} 	\end{subfigure} 	\begin{subfigure}[b]{0.32\linewidth} 		\centering 	\begin{tabular}{ c | c | c | c |} \cline{2-4} 		& Fruits & Center & Vol.
		\\ \hline\hline 
		\makecell[c]{Old\\(all)} & \makecell[r]{41.1\\ $\pm$ 3.3} & \makecell[r]{2.4\\ $\pm$ 0.4} & \makecell[r]{-0.19\\ $\pm$ 0.27} \\ \hline
		\makecell[c]{SE\\(all)} & \makecell[r]{40.9\\ $\pm$ 3.9} & \makecell[r]{1.9\\ $\pm$ 0.4} & \makecell[r]{0.22\\ $\pm$ 0.22} \\ \hline\hline
		\makecell[c]{Old\\(MA0)} & \makecell[r]{28.0\\ $\pm$ 4.7} & \makecell[r]{2.5\\ $\pm$ 0.5} & \makecell[r]{0.44\\ $\pm$ 0.06} \\ \hline
		\makecell[c]{SE\\(MA0)} & \makecell[r]{33.6\\ $\pm$ 3.9} & \makecell[r]{1.3\\ $\pm$ 0.3} & \makecell[r]{0.70\\ $\pm$ 0.04} \\ \hline\hline
		\makecell[c]{Old\\(MA50)} & \makecell[r]{12.4\\ $\pm$ 3.6} & \makecell[r]{2.1\\ $\pm$ 0.6} & \makecell[r]{0.71\\ $\pm$ 0.04} \\ \hline
		\makecell[c]{SE\\(MA50)} & \makecell[r]{27.2\\ $\pm$ 3.9} & \makecell[r]{1.2\\ $\pm$ 0.3} & \makecell[r]{0.80\\ $\pm$ 0.02} \\ \hline
	\end{tabular} \caption{Results Scenario 3} \label{tab:res_table_3} 	\end{subfigure} 	\caption{\textit{(a-c)}: Our three simulated scenarios.
	\textit{Scenario 1:} Simulated environment with static arm, as used in \cite{zaenker2020viewpoint}, with 4 plants and 14 fruits.
	\textit{Scenario 2}: Simulated environment with retractable arm, as in \cite{zaenker2020viewpoint}, with 4 plants and 28 fruits.
	\textit{Scenario 3}: New simulated environment.
It also uses the retractable arm, but contains a total of 12 plants, ordered in two rows, with different growth stages and fruits.
There are a total of 47 red fruits in this environment.\\ 	\textit{(d-f)}: Results for the three simulated scenarios.
	For each tested approach, 20 trials with a duration of three minutes each were performed.
	The plots show the average number of detected fruits of our superellipsoid matcher (\textbf{SE}) compared to our old octree-based approach (\textbf{Old}), with no minimum accuracy (\textbf{all}), a minimum accuracy of 0 (\textbf{MA0}), and a minimum accuracy of 0.5 (\textbf{MA50}).
	As can be seen, while the number of detected fruits with no accuracy bounds is similar, our new approach detects much more fruits with good accuracy.\\ 	\textit{(g-i)}: The average numerical results at the end of the episodes regarding number of detected fruits (\textbf{Fruits}), center distance in cm (\textbf{Center}), and volume accuracy (\textbf{Vol.}).
} 
	\label{fig:res_detected_fruits}
\end{figure*}

\Cref{fig:det_fruits_w14,fig:det_fruits_w19,fig:det_fruits_w22} illustrate the number of detected fruits over the three minutes planning time, and \Cref{tab:res_table_1,tab:res_table_2,tab:res_table_3} show the quantitative results for the three scenarios.

While the number of detected fruits with no volume accuracy bounds is similar to our old approach, our current approach with the superellipsoid matcher recognizes signifcantly greater number of fruits, when the minimum volume accuracy bound is applied.
With a minimum accuracy of 0.5, we detect an average of 7.7 fruits with our new aapproach compared to 4.7 with the old approach in Scenario 1, 17.7 compared to 6.0 in Scenario 2, and 27.2 compared to 12.4 in Scenario 3.

It can also be seen that with our old approach, the number of detected fruits with minimum accuracy drops towards the end.
This can happen if nearby fruits are combined to a single fruit, which greatly decreases the volume accuracy.
Since our new method uses a more accurate map and better clustering, this happens less often, leading to a considerable improvement in count and volume estimation.

The average center distances are similar for both approaches.
However, the average volume accuracy of our superellipsoid matcher is much higher.
When counting all detected fruits, with average accuracies between 0.22 and 0.35, it is still quite low.
This is because the clustering still can fail if there are not enough fruit points detected and nearby fruits are merged, leading to overestimated shapes with negative accuracies, which bring down the average.
When removing those outliers, which was done via prior assumptions about the maximum fruit size, the average accuracy increases appreciably, with accuracies between 0.67 and 0.70 compared to only between 0.35 and 0.45 with our old approach.
This accuracy covers on average over 70\% of the detected fruits in Scenario 1 and over 80\% in Scenario 2 and 3.

The evaluation demonstrates that our approach with superellipsoid based shape completion can lead to a higher number of accurately detected fruits.
While there are still some outliers, the volume accuracy for most of the sweet peppers shows that they can be satisfactorily approximated through superellipsoids, even if only partially observed by our viewpoint planner.

\subsection{Real-World Glasshouse Experiment}

To evaluate how our approach handles real-word data, we utilized a recording of the data feed gathered from our previously developed viewpoint planner in a capsicum glasshouse.
Color and depth images were generated using a RealSense L515 Lidar sensor, and the data includes the arm configuration and the camera poses.

The recorded images are passed through a neural network~\cite{halstead2020fruit} to generate fruit masks which are fused with the depth images to generate the fruit point clouds used for voxblox mapping.
In addition to the fruit map used to estimate shapes with superellipsoids, we generate a map from the full point cloud for visualization purposes.

\begin{figure}
	\centering
	\includegraphics[width=\linewidth]{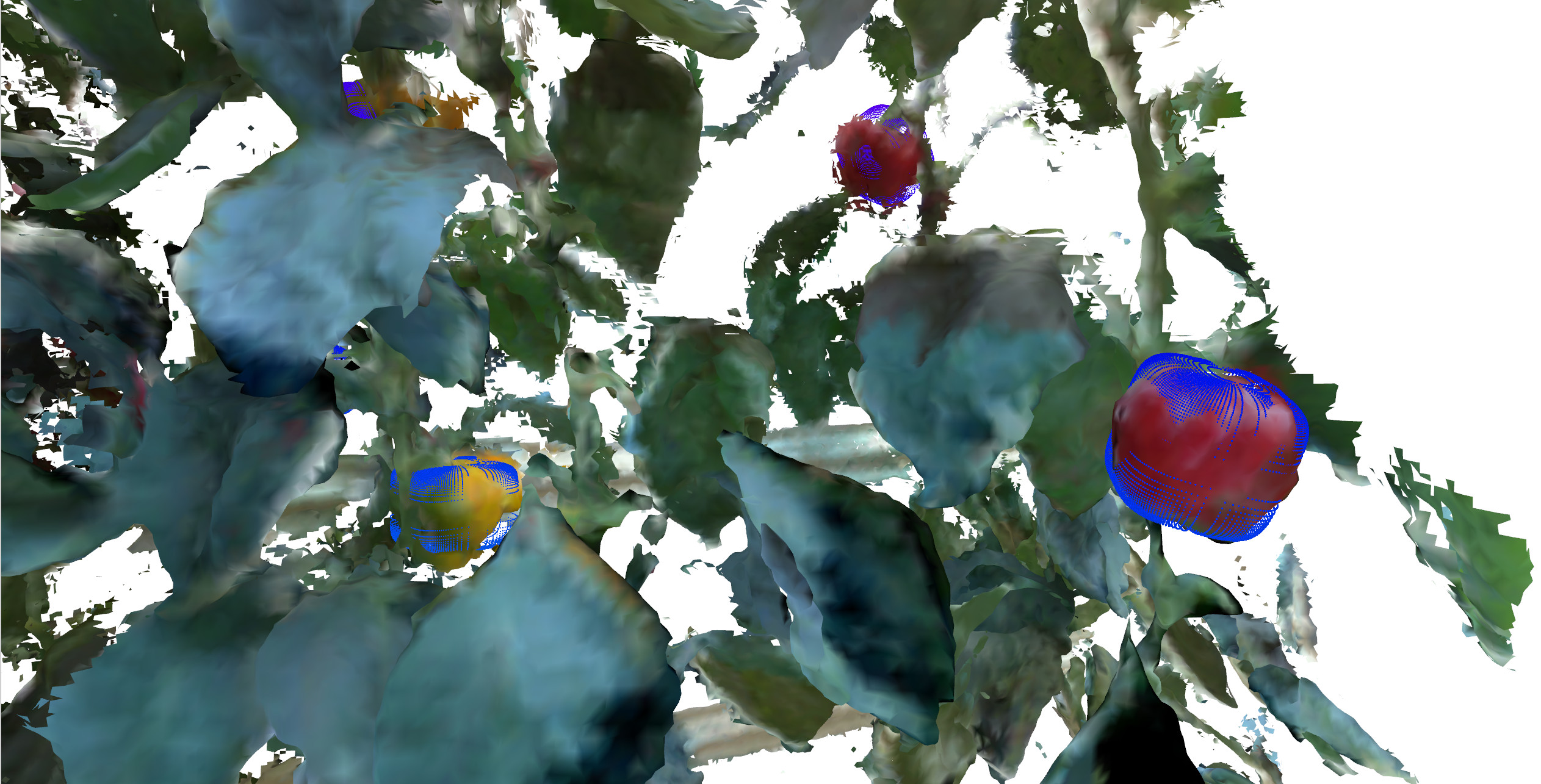}
	\caption{Generated map with superellipsoids. Blue point clouds represent the estimated surface of the matched shapes.}
	\label{fig:rw_image}
\end{figure}

\begin{figure}
	\centering
	\begin{subfigure}[b]{0.49\linewidth} 		\centering 		\includegraphics[width=\linewidth]{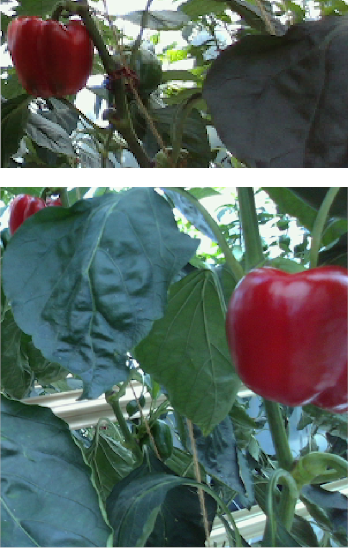} 		\caption{Captured photos}
		\label{fig:glasshouse_photo}
	\end{subfigure}
	\begin{subfigure}[b]{0.49\linewidth} 		\centering 		\includegraphics[width=\linewidth]{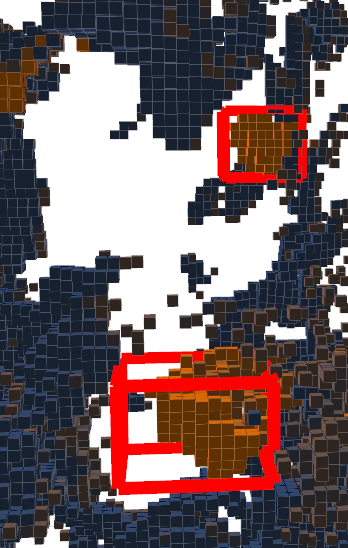} 		\caption{ROI octree} 
		\label{fig:glasshouse_rois}
	\end{subfigure}
	\caption{\textit{Left}: Captured RGB images of the glasshouse in the segment mapped in~\Cref{fig:rw_image} . \textit{Right}: The octree-based map generated in~\cite{zaenker2020viewpoint}, with red bounding boxes visualizing the matched fruit volume. The two fruits are the same as the two red fruits on the right in~\Cref{fig:rw_image}}
	\label{fig:old_results}
\end{figure}

In \Cref{fig:rw_image}, you can see the generated full TSDF map of a section of the glasshouse, with the superellipsoids matched from the fruit-only map visualized as blue point clouds.
For comparison, \Cref{fig:old_results} shows the results of the same segment using the octree-based map presented in \cite{zaenker2020viewpoint}.
It can be seen that our new method provides both a more accurate map and a better approximation of the fruit shapes.

\section{Conclusions}
\label{sec:concl}

To move a step closer to autonomous crop monitoring and yield estimation, we developed a framework to accurately estimate the shape of fruits such as sweet peppers.
Our system is embedded into a viewpoint planning approach and builds up a map of the fruits in form of truncated signed distance fields, on which fruits are clustered.
Subsequently, the shape of clustered fruits is approximated by fitting superellipsoids to the surface points.
As the evaluation with a simulated robotic arm equipped with an RGB-D camera demonstrates, our system can detect a majority of the fruits with good accuracy, and the qualitative results on real-world data show that it is applicable under realistic conditions in commercial glasshouse environments.

In the future, we plan to use the estimated shapes to determine the missing parts of partially detected fruits to guide our viewpoint planner to achieve better coverage.

\bibliographystyle{IEEEtran}
\bibliography{refs}

\begin{thebibliography}{10}
\providecommand{\url}[1]{#1}
\csname url@samestyle\endcsname
\providecommand{\newblock}{\relax}
\providecommand{\bibinfo}[2]{#2}
\providecommand{\BIBentrySTDinterwordspacing}{\spaceskip=0pt\relax}
\providecommand{\BIBentryALTinterwordstretchfactor}{4}
\providecommand{\BIBentryALTinterwordspacing}{\spaceskip=\fontdimen2\font plus
\BIBentryALTinterwordstretchfactor\fontdimen3\font minus
  \fontdimen4\font\relax}
\providecommand{\BIBforeignlanguage}[2]{{%
\expandafter\ifx\csname l@#1\endcsname\relax
\typeout{** WARNING: IEEEtran.bst: No hyphenation pattern has been}%
\typeout{** loaded for the language `#1'. Using the pattern for}%
\typeout{** the default language instead.}%
\else
\language=\csname l@#1\endcsname
\fi
#2}}
\providecommand{\BIBdecl}{\relax}
\BIBdecl

\bibitem{granier2021migrant}
M.-L. Aug{\`e}re-Granier, ``Migrant seasonal workers in the european
  agricultural sector,'' \emph{Briefing. EU: European Parliamentary Research
  Service. https://www. europarl. europa.
  eu/RegData/etudes/BRIE/2021/689347/EPRS\_BRI (2021) 689347\_ EN. pdf}, 2021.

\bibitem{rembold2013using}
F.~Rembold, C.~Atzberger, I.~Savin, and O.~Rojas, ``Using low resolution
  satellite imagery for yield prediction and yield anomaly detection,''
  \emph{Remote Sensing}, vol.~5, no.~4, pp. 1704--1733, 2013.

\bibitem{maimaitijiang2020crop}
M.~Maimaitijiang, V.~Sagan, P.~Sidike, A.~M. Daloye, H.~Erkbol, and F.~B.
  Fritschi, ``Crop monitoring using satellite/uav data fusion and machine
  learning,'' \emph{Remote Sensing}, vol.~12, no.~9, p. 1357, 2020.

\bibitem{stefas2019vision}
N.~Stefas, H.~Bayram, and V.~Isler, ``Vision-based monitoring of orchards with
  uavs,'' \emph{Computers and Electronics in Agriculture}, vol. 163, p. 104814,
  2019.

\bibitem{qiao2005mapping}
J.~Qiao, A.~Sasao, S.~Shibusawa, N.~Kondo, and E.~Morimoto, ``Mapping yield and
  quality using the mobile fruit grading robot,'' \emph{Biosystems
  Engineering}, vol.~90, no.~2, pp. 135--142, 2005.

\bibitem{shafiekhani2017vinobot}
A.~Shafiekhani, S.~Kadam, F.~B. Fritschi, and G.~N. DeSouza, ``Vinobot and
  vinoculer: two robotic platforms for high-throughput field phenotyping,''
  \emph{Sensors}, vol.~17, no.~1, p. 214, 2017.

\bibitem{lehnert2016sweet}
C.~Lehnert, I.~Sa, C.~McCool, B.~Upcroft, and T.~Perez, ``Sweet pepper pose
  detection and grasping for automated crop harvesting,'' in \emph{Proc.~of the
  IEEE Intl.~Conf.~on Robotics \& Automation (ICRA)}.\hskip 1em plus 0.5em
  minus 0.4em\relax IEEE, 2016, pp. 2428--2434.

\bibitem{zaenker2020viewpoint}
T.~Zaenker, C.~Smitt, C.~McCool, and M.~Bennewitz, ``Viewpoint planning for
  fruit size and position estimation,'' in \emph{Proc.~of the IEEE/RSJ
  Intl.~Conf.~on Intelligent Robots and Systems (IROS)}, 2021.

\bibitem{oleynikova2017voxblox}
H.~Oleynikova, Z.~Taylor, M.~Fehr, R.~Siegwart, and J.~Nieto, ``Voxblox:
  Incremental 3d euclidean signed distance fields for on-board mav planning,''
  in \emph{Proc.~of the IEEE/RSJ Intl.~Conf.~on Intelligent Robots and Systems
  (IROS)}, 2017.

\bibitem{wang2020fruit}
Y.~Wang and Y.~Chen, ``Fruit morphological measurement based on
  three-dimensional reconstruction,'' \emph{Agronomy}, vol.~10, no.~4, p. 455,
  2020.

\bibitem{jadhav2019volumetric}
T.~Jadhav, K.~Singh, and A.~Abhyankar, ``Volumetric estimation using 3d
  reconstruction method for grading of fruits,'' \emph{Multimedia Tools and
  Applications}, vol.~78, no.~2, pp. 1613--1634, 2019.

\bibitem{roy2017active}
P.~Roy and V.~Isler, ``Active view planning for counting apples in orchards,''
  in \emph{2017 IEEE/RSJ International Conference on Intelligent Robots and
  Systems (IROS)}.\hskip 1em plus 0.5em minus 0.4em\relax IEEE, 2017, pp.
  6027--6032.

\bibitem{kang2020visual}
H.~Kang, H.~Zhou, and C.~Chen, ``Visual perception and modeling for autonomous
  apple harvesting,'' \emph{IEEE Access}, vol.~8, pp. 62\,151--62\,163, 2020.

\bibitem{kang2020fruit}
H.~Kang and C.~Chen, ``Fruit detection, segmentation and 3d visualisation of
  environments in apple orchards,'' \emph{Computers and Electronics in
  Agriculture}, vol. 171, p. 105302, 2020.

\bibitem{lin2019guava}
G.~Lin, Y.~Tang, X.~Zou, J.~Xiong, and J.~Li, ``Guava detection and pose
  estimation using a low-cost rgb-d sensor in the field,'' \emph{Sensors},
  vol.~19, no.~2, p. 428, 2019.

\bibitem{ge2020symmetry}
Y.~Ge, Y.~Xiong, and P.~J. From, ``Symmetry-based 3d shape completion for fruit
  localisation for harvesting robots,'' \emph{biosystems engineering}, vol.
  197, pp. 188--202, 2020.

\bibitem{thrun2005shape}
S.~Thrun and B.~Wegbreit, ``Shape from symmetry,'' in \emph{Tenth IEEE
  International Conference on Computer Vision (ICCV'05) Volume 1},
  vol.~2.\hskip 1em plus 0.5em minus 0.4em\relax IEEE, 2005, pp. 1824--1831.

\bibitem{gongal2018apple}
A.~Gongal, M.~Karkee, and S.~Amatya, ``Apple fruit size estimation using a 3d
  machine vision system,'' \emph{Information Processing in Agriculture},
  vol.~5, no.~4, pp. 498--503, 2018.

\bibitem{wang2017tree}
Z.~Wang, K.~B. Walsh, and B.~Verma, ``On-tree mango fruit size estimation using
  rgb-d images,'' \emph{Sensors}, vol.~17, no.~12, p. 2738, 2017.

\bibitem{pirovano2012kinfu}
M.~Pirovano, ``Kinfu--an open source implementation of kinect fusion+ case
  study: implementing a 3d scanner with pcl,'' \emph{UniMi, Tech. Rep.}, 2012.

\bibitem{zaenker2020ecmr}
T.~Zaenker, C.~Lehnert, C.~McCool, and M.~Bennewitz, ``Combining local and
  global viewpoint planning for fruit coverage,'' in \emph{Proc.~of the
  Europ.~Conf.~on Mobile Robotics (ECMR)}, 2021.

\bibitem{halstead2020fruit}
M.~Halstead, S.~Denman, F.~Clinton, and C.~McCool, ``Fruit detection in the
  wild: The impact of varying conditions and cultivar,'' in \emph{Digital Image
  Computing: Techniques and Applications (DICTA)}, 2020.

\bibitem{traa2013least}
J.~Traa, ``Least-squares intersection of lines,'' \emph{University of Illinois
  Urbana-Champaign (UIUC)}, 2013.

\bibitem{jaklic2000superquadrics}
\BIBentryALTinterwordspacing
A.~Jakli{\v{c}}, A.~Leonardis, and F.~Solina, \emph{Superquadrics and Their
  Geometric Properties}.\hskip 1em plus 0.5em minus 0.4em\relax Dordrecht:
  Springer Netherlands, 2000, pp. 13--39. [Online]. Available:
  \url{https://doi.org/10.1007/978-94-015-9456-1_2}
\BIBentrySTDinterwordspacing

\bibitem{ceres-solver}
S.~Agarwal, K.~Mierle, and Others, ``Ceres solver,''
  \url{http://ceres-solver.org}.

\end{thebibliography}

\end{document}